\documentclass{article}
\pdfoutput=1
\newcommand{\norm}[1]{\left\lVert#1\right\rVert}

\newcommand\myeqb{\stackrel{\mathclap{\normalfont\mbox{\text{\small{(1)}}}}}{=}}
\newcommand\myeqc{\stackrel{\mathclap{\normalfont\mbox{\text{\small{(2)}}}}}{\leq}}
\newcommand\myeqd{\stackrel{\mathclap{\normalfont\mbox{\text{\small{(3)}}}}}{=}}

\usepackage[nonatbib,preprint]{neurips_2020}
\usepackage{booktabs, bookmark, units, RobStd, amsmath} 
\usepackage[utf8]{inputenc} 
\usepackage[T1]{fontenc}    
\usepackage{hyperref}       
\usepackage{url}            
\usepackage{booktabs}       
\usepackage{amsfonts}       
\usepackage{nicefrac}       
\usepackage{microtype}      
\usepackage{threeparttable, algorithm, algorithmicx}
\usepackage[noend]{algpseudocode}
\usepackage{caption, subcaption, units} 
\usepackage{wrapfig}
\usepackage{enumitem,kantlipsum}
\usepackage{mathtools}

\usepackage{multirow}
\usepackage[table,xcdraw]{xcolor}

\title{OrthoReg: Robust Network Pruning Using Orthonormality Regularization}
\author{%
  Ekdeep Singh~Lubana\\
  \texttt{eslubana@umich.edu} \\
  \And
  Puja Trivedi \\
  \texttt{pujat@umich.edu} \\
  \AND
  Conrad Hougen \\
  \texttt{chougen@umich.edu} \\
  \And
  Robert P.\ Dick \\
  \texttt{dickrp@umich.edu} \\
  \And
  Alfred O.\ Hero \\
  \texttt{hero@eecs.umich.edu} \\
  \AND
  \vspace{-15pt}\\
  EECS Department\\
  University of Michigan\\
  Ann Arbor, MI 48105 \\
  }

\begin{document}

\maketitle

\begin{abstract}
Network pruning in Convolutional Neural Networks (CNNs) has been extensively investigated in recent years. To determine the impact of pruning a group of filters on a network's accuracy, state-of-the-art pruning methods consistently assume filters of a CNN are independent. This allows the importance of a group of filters to be estimated as the sum of importances of individual filters. However, overparameterization in modern networks results in highly correlated filters that invalidate this assumption, thereby resulting in incorrect importance estimates. To address this issue, we propose OrthoReg, a principled regularization strategy that enforces orthonormality on a network's filters to reduce inter-filter correlation, thereby allowing reliable, efficient determination of group importance estimates, improved trainability of pruned networks, and efficient, simultaneous pruning of large groups of filters. When used for iterative pruning on VGG-13, MobileNet-V1, and ResNet-34, OrthoReg consistently outperforms five baseline techniques, including the state-of-the-art, on CIFAR-100 and Tiny-ImageNet. For the recently proposed Early-Bird Ticket hypothesis, which claims networks become amenable to pruning early-on in training and can be pruned after a few epochs to minimize training expenditure, we find OrthoReg significantly outperforms prior work. Code available at \url{https://github.com/EkdeepSLubana/OrthoReg}.
\end{abstract}

\section{Introduction}
Convolutional Neural Networks (CNNs) have achieved state-of-the-art (SOTA) performance on several computer vision applications. However, their heavy computational requirements impede deployment in resource-constrained scenarios. Network pruning has been extensively studied as a solution to this problem, wherein unimportant filters are removed to reduce computational requirements while maintaining application performance.

\begin{wrapfigure}{R}{6cm}
\centering
	\includegraphics[width=\linewidth]{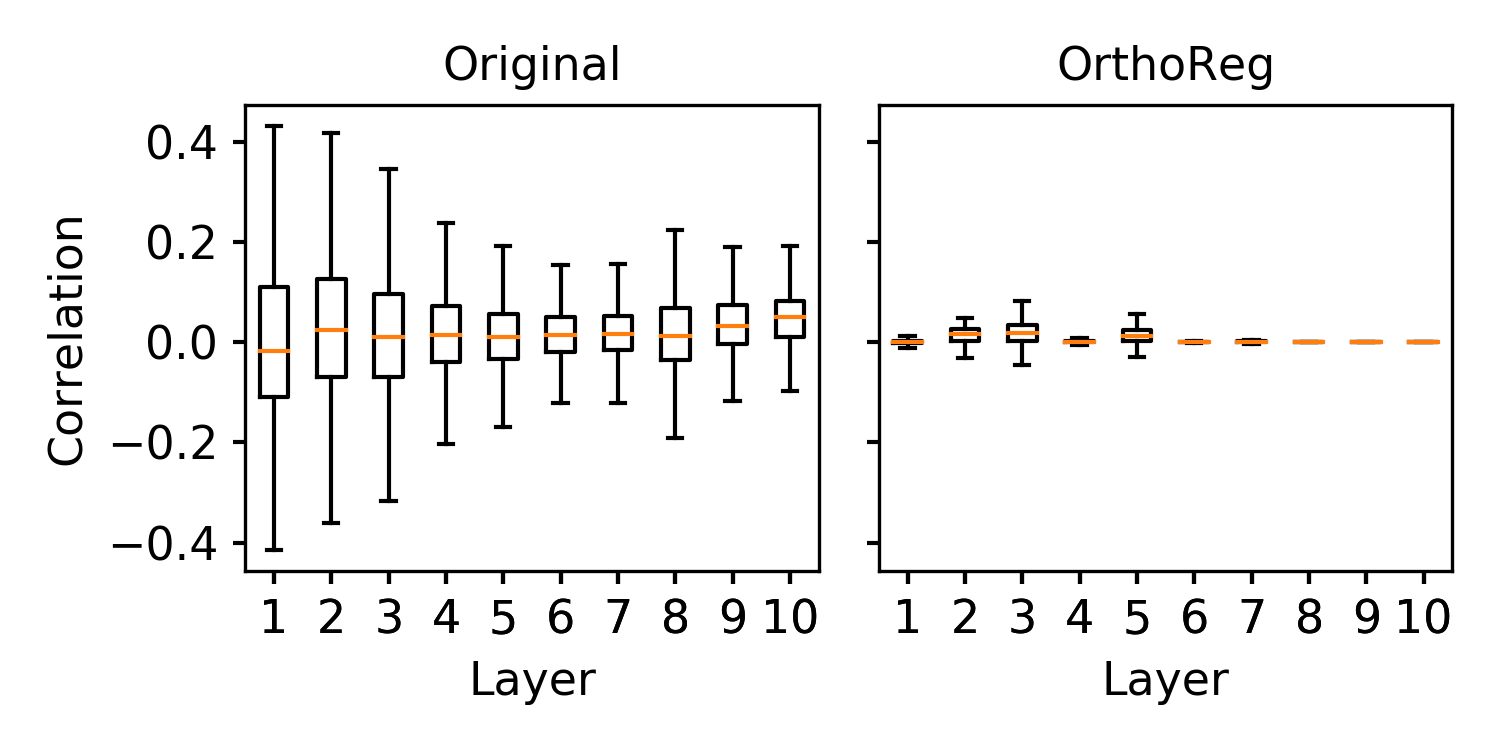}
	\caption{Layerwise partial correlation statistics between filters of a VGG-13 model and its OrthoReg version. Partial correlation~\cite{parcorr} is a measure of operational redundancy that indicates high predictability of one filter in a layer from the others in that layer. We find filters in the original model have much higher partial correlation than their OrthoReg counterparts.}
	\label{fig:intromap}
\end{wrapfigure}

Accurately estimating importance is essential for pruning algorithms to achieve effective combinations of accuracy and computational efficiency. Prior work determine the importance of a filter by estimating the impact of pruning that filter on a network's loss. These works~\cite{nvidia, fisher, tfo, rdt} consistently assume the filters of a network are independent, which allows the \emph{joint} impact of pruning a group of filters, i.e., the importance of a group, to be calculated as the sum of \emph{individual} impacts of pruning each filter in it. However, overparameterization in modern CNNs results in high inter-filter correlation (see \autoref{fig:intromap}) and invalidates the independent filter assumption. In particular, for recent SOTA importance metrics~\cite{nvidia, fisher}, we find that while the importance of an individual filter can be reliably estimated, estimates for the importance of a group are highly unreliable--i.e, pruning a large number of filters using these methods results in loss of important filters and deteriorates trainability of pruned networks~\cite{signalprop}, severely impacting accuracy and rendering the ability to regain lost accuracy via retraining ineffective. To mitigate this, prior work use a slow, iterative process that cycles between pruning only one~\cite{fisher} or a few (e.g., 2\%~\cite{nvidia}) filters at a time, recomputing importance, and retraining the pruned network after each round. This makes network pruning a highly expensive process--e.g., pruning ResNet-34 by 84\%, at a rate of 2\% filters per round, requires 42 rounds! 

\begin{figure*}
\centering
\includegraphics[width=\linewidth]{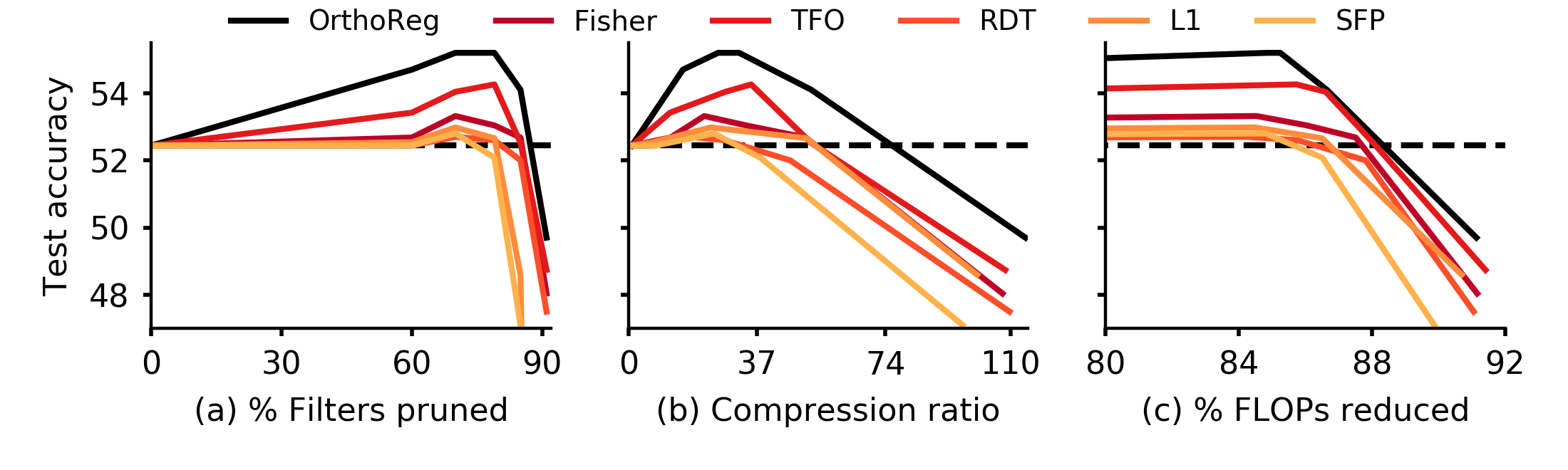}
\caption{ResNet-34 pruning on Tiny-ImageNet. Methods are evaluated with respect to (a) \% Filters pruned, (b) Compression ratio, and (c) \% FLOPs reduced. Dotted line shows accuracy of the uncompressed model. We evaluate our method against 5 baselines: Fisher~\cite{nvidia, fisher}, TFO~\cite{tfo}, RDT~\cite{rdt}, L1~\cite{l1}, and SFP~\cite{soft}.}
\label{fig:res34}
\end{figure*}

To ameliorate this situation, it is important to design techniques that enable several filters to be removed in a single round. We thus propose OrthoReg, a regularization objective that enforces orthonormality on filters in a CNN, which, as we show (see \autoref{sec:orthoreg}), allows importance estimates for a group of filters to be reliably approximated as the sum of their individual importances and improves the trainability of pruned networks by ensuring better preservation of error norm during backward propagation. During initial rounds of pruning, when redundancy (i.e., inter-filter correlation) is high, better importance estimates enable us to remove a large number of filters; later, when redundancy reduces, better trainability allows lost accuracy to be regained via retraining, thus ensuring pruning continues to be possible. Together, these benefits make OrthoReg highly \emph{robust}--i.e, without loss in accuracy, OrthoReg is able to prune several filters (e.g., 40--60\%) in a single round. For iterative pruning, OrthoReg significantly outperforms 5 baselines for 3 different models trained on CIFAR-100 and Tiny-ImageNet. For example, in just 3 rounds, without loss in accuracy, OrthoReg prunes a ResNet-34 model trained on Tiny-ImageNet by 84\%, while improving the compression ratio\footnote{Compression Ratio is defined in this paper as size of original model / size of pruned model.} by 1.43$\times$ over the SOTA method~\cite{nvidia, fisher} at an equivalent reduction in FLOPs (see \autoref{fig:res34}). For the recently proposed Early-Bird Ticket hypothesis, which claims networks become amenable to pruning early-on in training and can be pruned after a few epochs to minimize training expenditure, we find that OrthoReg consistently produces better models than prior work~\cite{ebt}. For example, without loss in accuracy, OrthoReg extracted Early-Bird Tickets on ResNet-34 achieve 7$\times$ higher compression ratio and 46\% higher reduction in FLOPs than other methods (see \autoref{tab:EBT}).

This paper is organized as follows: In \autoref{sec:corr}, we demonstrate the independent filter assumption in SOTA pruning methods results in inaccurate importance estimates. In \autoref{sec:orthoreg}, we describe OrthoReg, a ``pruning-aware'' training framework that improves the validity of the independent filter assumption and improves pruned networks' trainability. In \autoref{sec:exptts}, we compare OrthoReg for iterative pruning with 5 baselines~\cite{nvidia, tfo, fisher, rdt, l1, soft} and test OrthoReg's single-shot pruning abilities by extracting Early-Bird Tickets~\cite{ebt} for several models.

\section{Related Work}
\label{sec:related}
Several pruning techniques have been recently proposed, wherein the general idea is to design ``importance'' metrics that estimate the impact of removing a particular filter in a CNN. The net impact of removing several such filters is then calculated as the sum of their individual importances. 

Recent interest in network pruning was reignited when Han et al.~\cite{han} showed that at minimal to no loss in accuracy, high orders of compression can be achieved by removing connections with small weight magnitude. This form of pruning, called \emph{unstructured} pruning, results in sparse filters that require specialized hardware to achieve useful energy or time reductions in practical systems~\cite{eie}. To enable efficient implementation on commodity hardware, \emph{structured} pruning methods remove entire filters instead of individual weights. Magnitude-based methods exist for structured pruning as well~\cite{l1, soft}. Liu et al.~\cite{netslim} re-imagine BatchNorm's scale parameters as gating variables that represent importance of a filter. Note that these methods were originally proposed for \emph{local} pruning--i.e., they estimate the importance of a filter relative to other filters in the same layer, thus requiring the user to specify a threshold on the number of filters to be pruned at each layer.

In contrast, several methods estimate a \emph{global} filter importance and require only a single, global threshold to decide which filters are to be pruned in a network. Such global pruning methods have been empirically shown to result in better post-pruning performance~\cite{shrinkbench}. The earliest works (e.g., OBD~\cite{obd} and OBS~\cite{obs}) on global pruning defined the ground truth for the importance of a parameter as the change in loss before and after its removal, with the overall goal of preserving network accuracy by preserving its loss. These works used a second-order Taylor expansion to approximate the change in loss; however, as computing a Hessian is expensive, such methods are impractical for modern neural networks. Interestingly, Molchanov et al.~\cite{tfo} recently showed that the first-order Taylor expansion of network loss is sufficient for computing high quality importance estimates. Since gradients are already available during the training process, this method is particularly efficient for large architectures. Recent works have extended this first-order Taylor formulation to demonstrate its link with Fisher information~\cite{nvidia, fisher} and Rate-Distortion Theory~\cite{rdt}, consequently deriving better estimates of impact of filter removal on a network's loss. Due to its better performance and higher computational efficiency, we discuss global structured pruning in this paper.

\section{Importance Estimation in CNNs}
\label{sec:corr}
Existing pruning techniques formulate the net importance of a combination of filters as the sum of their individual importances. Therefore, these methods, either implicitly~\cite{l1, soft} or explicitly~\cite{nvidia, tfo, fisher}, rely on CNN filters being independent. However, several recent works have demonstrated that the extracted representations within a layer of a neural network are highly correlated, with only inter-layer representations being relatively dissimilar~\cite{svcca, pwcca, cka}. These works therefore invalidate the underlying independent filter assumption in prior pruning methods. To analyze the impact of inter-filter correlation in designing importance estimates for a group of filters, we investigate global pruning techniques based on loss preservation. These methods have led to recent SOTA results~\cite{nvidia, fisher}.

Consider the following method for defining the importance of a filter of $w \times h$ dimensions and $c_{in}$ input channels, parametrized by $\mathbf{w_{i}} \in {\rm I\!R}^{whc_{in} \times 1}$, as a function of its influence on the loss function
\begin{equation}
\label{eqn:base_imp}
I\left(\mathbf{w_{i}}\right) = \left(L\left(\Theta\right) - L\left(\Theta, \mathbf{w_{i}} = 0\right)\right)^{2},
\end{equation}
where $\Theta$ denotes the parameters of a CNN and $L\left(\Theta\right)$ denotes its loss. Using Taylor's first-order expansion about $\mathbf{w_{i}} = \mathbf{0}$ yields
\begin{equation}
\label{eqn:ind_imp}
I\left(\mathbf{w_{i}}\right) = \left(\mathbf{w_{i}}^{T} \nabla_{\mathbf{w_{i}}} L \right)^{2},
\end{equation}
where $L\left(\Theta\right)$ is denoted as $L$ for brevity. Squaring the change in loss ensures that the final metric better focuses on preserving the distribution of the original network's output. As has been shown before, this is more valuable from a compression standpoint~\cite{rdt, tfo, nvidia}.

\autoref{eqn:ind_imp} denotes the impact of pruning an individual filter on the network's overall loss. Since multiple filters are pruned in a single round, an importance metric for groups of filters is needed. To this end, prior work approximate the net importance of a group of filters as the arithmetic sum of the individual importances of filters contained within it. For example, if we are pruning a group of N filters $\left[\mathbf{W}\right] = \{\mathbf{w_{1}}, \mathbf{w_{2}}, \dots \mathbf{w_{N}}\}$, then prior work calculate the net importance of this group as $I\left(\left[\mathbf{W}\right]\right) = \sum_{i = 1}^{N} \left(\mathbf{w_{i}}^{T} \nabla_{\mathbf{w_{i}}} L \right)^{2}$. However, using a Taylor's first-order expansion about $\left[\mathbf{W}\right] = \{\mathbf{0}, \mathbf{0}, \dots \mathbf{0}\}$, the net impact of pruning the group is estimated as
\begin{equation}
\begin{split}
\label{eqn:net_imp}
I\left(\left[\mathbf{W}\right]\right) & = \left(\sum_{i = 1}^{N} \mathbf{w_{i}}^{T} \nabla_{\mathbf{w_{i}}} L \right)^{2} \\
&= \sum_{i = 1}^{N} \left(\mathbf{w_{i}}^{T} \nabla_{\mathbf{w_{i}}} L \right)^{2} + 2 \sum_{i = 1}^{N} \sum_{j = i+1}^{N} \left(\mathbf{w_{i}}^{T} \nabla_{\mathbf{w_{i}}} L \right) \left(\mathbf{w_{j}}^{T} \nabla_{\mathbf{w_{j}}} L \right).
\end{split}
\end{equation}
The term $\left(\mathbf{w_{i}}^{T} \nabla_{\mathbf{w_{i}}} L \right) \left(\mathbf{w_{j}}^{T} \nabla_{\mathbf{w_{j}}} L \right) = \nabla_{\mathbf{w_{i}}} L^{T} \left( \mathbf{w_{i}} \mathbf{w_{j}}^{T} \right) \nabla_{\mathbf{w_{j}}} L$ results in biased importance estimates and can be interpreted as the joint impact of simultaneously pruning filters $\mathbf{w_{i}}$ and $\mathbf{w_{j}}$ due to their correlated nature. We call $\sum_{i = 1}^{N} \sum_{j = i+1}^{N} \left(\mathbf{w_{i}}^{T} \nabla_{\mathbf{w_{i}}} L \right) \left(\mathbf{w_{j}}^{T} \nabla_{\mathbf{w_{j}}} L \right)$ the ``correlation term''. 

By focusing on an individual filter's impact on network loss, prior work neglect this correlation term. If this joint impact of pruning several filters is considered, searching for the least important group of $N$ filters requires an $N$-way search over all filters to find a group that minimizes $|\sum_{i = 1}^{N} \mathbf{w_{i}}^{T} \nabla_{\mathbf{w_{i}}} L|$. To circumvent this problem, which has a computational complexity increasing exponentially in $N$, prior work use a slow, iterative process that cycles between pruning only one~\cite{fisher} or a few (e.g., 2\%~\cite{nvidia}) filters at a time, recomputing importance, and retraining after each round. This makes pruning an expensive process and necessitates design of methods capable of removing a large numbers of filters in a single round.

\begin{algorithm}[t]
\caption{OrthoReg: Pruning Networks Trained With Orthonormality Regularization}
\label{training-algo}
\begin{algorithmic}[1]
\State {\bf input} $net_{base}$: base network; $N$: \# of pruning rounds; ($p_{1}$, $p_{2}$, $\cdots$, $p_{N}$): percentage of filters to be pruned at a given round;  
\State $net_{0}$ = Fine-tune $net_{base}$ with orthonormality regularized loss (\autoref{eqn:loss_total})
\For{k = 1:N}
			\State $net_{k}$ = prune $p_{k}\%$ least important filters from $net_{k-1}$
			\If{k \textless N} \State Retrain $net_{k}$ with orthonormality regularized loss
			\Else \State Retrain $net_{N}$ without orthonormality regularized loss
			\EndIf
		\EndFor
\State Return $net_{N}$
\end{algorithmic}
\end{algorithm}

\section{OrthoReg}
\label{sec:orthoreg}

As shown in \autoref{sec:corr}, assuming independent filters results in a biased, inaccurate importance estimate, and determining the exact metric is computationally prohibitive. However, if the correlation term were zero, the importance estimate of an $N$-filter group could be directly calculated as the arithmetic sum of the importances of the individual filters contained within it. We therefore formulate a regularization objective that minimizes the correlation term. 

First, recall that $\left(\mathbf{w_{i}}^{T} \nabla_{\mathbf{w_{i}}} L \right) \left(\mathbf{w_{j}}^{T} \nabla_{\mathbf{w_{j}}} L \right) = \nabla_{\mathbf{w_{i}}} L^{T} \left( \mathbf{w_{i}} \mathbf{w_{j}}^{T} \right) \nabla_{\mathbf{w_{j}}} L$. Clearly, minimizing the outer product of each $\left(i, j\right)^{th}$ pair of filters will minimize the correlation term, thereby producing unbiased estimates for the importance of a group of filters. In the following, we only discuss how to minimize the correlation term corresponding to filters belonging to the same layer. Inter-layer filters generally produce dissimilar representations~\cite{svcca, pwcca, cka}, so the independent filter assumption is valid for them.

Consider a network with $L$ layers, such that the $l^{th}$ layer has a weight matrix denoted by $\mathbf{W}(l) \in {\rm I\!R}^{whc_{in} \times M_{l}}$, where $M_{l}$ is the number of filters in layer $l$. For pairs of filter outer products, we have
\begin{equation}
\label{eqn:deriv}
\begin{split}
&\norm{\sum_{i=1}^{M_{l}} \sum_{j=1}^{M_{l}} \mathbf{w_{i}}(l) \mathbf{w_{j}}(l)^{T}}_{F} = \norm{\sum_{i=1}^{M_{l}} \mathbf{w_{i}}(l) \sum_{j=1}^{M_{l}} \mathbf{w_{j}}(l)^{T}}_{F} \myeqb \norm{\sum_{i=1}^{M_{l}} \sum_{j=1}^{M_{l}} \mathbf{w_{i}}(l)^{T} \mathbf{w_{j}}(l)}_{F} \\
&\myeqc \sum_{i=1}^{M_{l}} \sum_{j=1}^{M_{l}} \norm{\mathbf{w_{i}}(l)^{T} \mathbf{w_{j}}(l)}_{F} \myeqd \sum_{i=1}^{M_{l}} \sum_{j=1}^{M_{l}} |\mathbf{w_{i}}(l)^{T} \mathbf{w_{j}}(l)| = \norm{\mathbf{W}(l)^{T} \mathbf{W}(l)}_{1},
\end{split}
\end{equation} 
where we used the following: (1) $\norm{\mathbf{A}\mathbf{A}^{T}}_{F} = \norm{\mathbf{A}^{T}\mathbf{A}}_{F}$ for any matrix $\mathbf{A}$; (2) the triangle inequality for norms; and (3) $\mathbf{w_{i}}(l)^{T} \mathbf{w_{j}}(l)$ is a scalar for all $(i, j)$. \autoref{eqn:deriv} suggests that if filters are orthogonal, the pairs of outer products are reduced, consequently minimizing the correlation term and allowing reliable approximation of group importance estimates as the sum of importance estimates for individual filters in that group. Motivated by this, we define the following regularizer
\begin{equation}
\label{eqn:reg}
\mathcal{L}_{\text{ortho}} = \sum_{l = 1}^{L} \alpha(l) \norm{\mathbf{W}(l)^{T} \mathbf{W}(l) - \mathbf{I}}_{1},
\end{equation}
where the identity matrix $\mathbf{I}$ is used to enforce the constraint $j \neq i$ and the coefficient $\alpha(l) = \frac{\sqrt{M_{l}}}{\sum_{l=1}^{L}\sqrt{M_{l}}}$ is used to lay greater emphasis on layers with more filters. Our overall learning objective is:
\begin{equation}
\label{eqn:loss_total}
\mathcal{L}_{\text{OrthoReg}} = \mathcal{L}_{\mathit{\text{cross entropy}}} + \lambda \mathcal{L}_{\text{ortho}},
\end{equation}
where $\mathcal{L}_{\mathit{\text{cross entropy}}}$ is the cross entropy loss, $\mathcal{L}_{\text{ortho}}$ is as defined in \autoref{eqn:reg}, and $\lambda$ is a regularization parameter. In practice, we find that $\lambda \in [0.001, 0.1]$ results in good performance.

\autoref{training-algo} describes our overall pruning method, named OrthoReg, in detail. In brief, OrthoReg involves fine-tuning a pretrained base network with the regularized loss (\autoref{eqn:loss_total}) to restructure its filters. The network is then pruned and retrained. To make an intermediate pruned network amenable to further pruning, retraining takes place with the regularizer enabled. The regularizer is disabled in the last round because no further pruning is to take place.

\subsection{Design Choices in OrthoReg: Orthonormality vs.\ Orthogonality}
Saxe et al.~\cite{saxe} show that initialization of linear networks with properly scaled orthogonal filters results in ``Dynamical Isometry'', a desirable property induced when the spectrum of a network's input-output Jacobian is concentrated around 1. This property ensures that the norm of the error signal is better preserved during backward propagation, thus improving network trainability. Recently, Lee at al.\ proposed SNIP~\cite{snip}, a method to perform single-shot, \emph{unstructured} pruning at initialization using a normalized form of \autoref{eqn:ind_imp}. In order to make \emph{post pruning} networks more amenable to training, they propose to induce ``Layerwise Dynamical Isometry'' by enforcing approximate orthonormality on remaining weights before training~\cite{signalprop}. This forces the spectrum for layerwise Jacobian (and hence the input-output Jacobian) of a network to concentrate around 1, thereby improving its trainability. 

We extend these observations for \emph{structured} pruning of networks. In particular, after a few rounds of pruning, removing even a small number of filters results in substantial loss of accuracy. This makes retraining crucial. However, extreme pruning (e.g., removing 75\% filters) severely deteriorates network trainability~\cite{signalprop}, rendering retraining ineffective. In order to address the same, we formulate OrthoReg to promote orthonormality\footnote{Enforcing orthonormality does not affect network output because BatchNorm's scale parameters adequately adapt themselves to perform any necessary scaling of filter norm.}, instead of mere orthogonality (see \autoref{eqn:reg}). By incentivizing orthonormal filters, OrthoReg pushes singular values of weight matrices in its networks to concentrate around 1. After pruning, filters in OrthoReg-pruned networks tend to remain close to orthonormal, resulting in an implicit, approximate Layerwise Dynamical Isometry that significantly improves their trainability. This allows OrthoReg to better utilize retraining for regaining any lost accuracy, thus enabling extreme pruning regimes as well (see \autoref{tab:EBT}). 

To summarize, OrthoReg uses \emph{orthogonality} to ensure unbiased importance estimates can be calculated for groups of filters, while using \emph{orthonormality} to improve network trainability for enabling extreme pruning regimes. These design choices also allow OrthoReg to significantly increase the number of filters that can be pruned in a single round, thus improving the ultimate accuracy and/or efficiency of its pruned models. We would like to note that in contrast to SNIP, OrthoReg is designed to perform \emph{structured} pruning and enforces orthonormality \emph{before} pruning, thus enabling calculation of reliable, unbiased importance estimates for groups of filters.

\section{Experiments}
\label{sec:exptts}
We perform three sets of experiments: (i) To analyze OrthoReg's effects on group importance estimates; (ii) To determine how robust OrthoReg is to pruning of a large number of filters; and (iii) To evaluate OrthoReg's effectiveness in an iterative pruning setting.  
\subsection{Correlation With Change in Loss: Estimator Reliability}
\begin{wrapfigure}{R}{6cm}
\centering
	\includegraphics[width=\linewidth, scale=0.7]{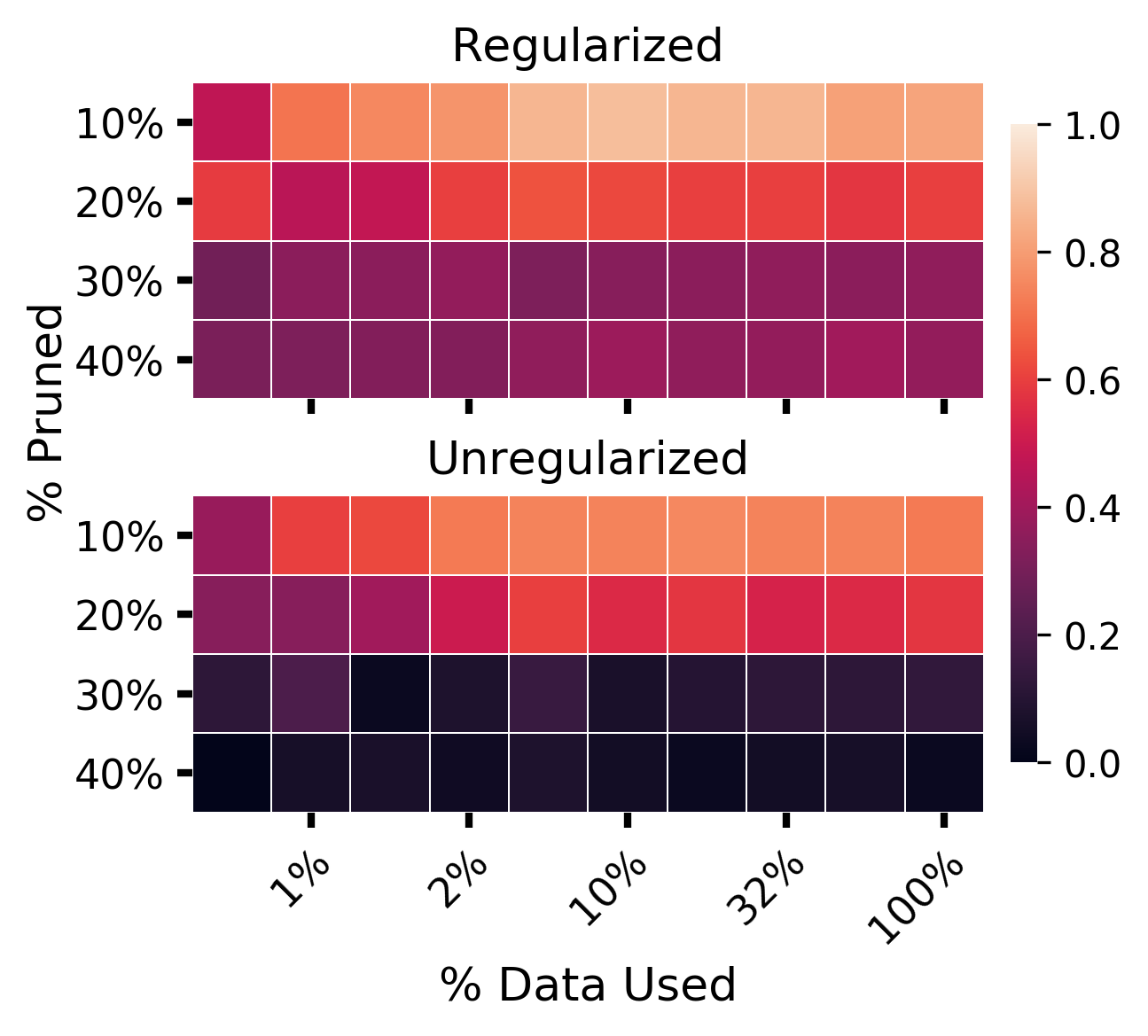}
	\caption{Correlation of estimated importance of groups of filters with squared change in loss for the Regularized and Unregularized models. X-axis shows \% training data used for estimating importance; Y-axis shows the size of a group relative to the total number of filters in the network.}
	\label{fig:corrgraph}
\end{wrapfigure}

We first investigate whether orthonormality regularization improves reliability of group importance estimates by evaluating the correlation of estimated importance of a group of filters with the defined baseline. In particular, we train a fully convolutional version of VGG-13 on CIFAR-100, retrain it with the orthonormality regularized objective, and prune a group of randomly selected filters from both the regularized and unregularized networks. We vary the group size from 10\% to 40\% of the total number of filters in the network. The baseline impact of pruning a group is calculated as the square of change in the network's loss before and after pruning. The estimated group importance is calculated as the sum of individual importance estimates of the filters contained in that group, where an individual filter's importance estimate is defined in \autoref{eqn:ind_imp}. Results from 100 trials were used to calculate correlation. \autoref{fig:corrgraph} shows the correlation between the baseline importance and estimated importance as a function of percentage of training data used for calculating the gradient in \autoref{eqn:ind_imp}. The OrthoReg model's correlation with squared change in loss is significantly higher than the unregularized version. Specifically, as more filters are pruned, OrthoReg continues to achieve high correlation (high reliability), while the unregularized model achieves near-zero correlation (low reliability). Therefore, using OrthoReg's more reliable importance estimates, we find that larger per-round pruning ratios (20--40\%) can be used.

\begin{table}[]
\centering
\caption{\label{tab:EBT}%
Test accuracies of Early-Bird Tickets extracted from models trained on CIFAR-100. Baseline accuracies are reported in parentheses. OrthoReg consistently outperforms other methods, allowing 40--60\% pruning in a single round. Results are averaged over 3 seeds. Compression ratios and \%FLOPs reduction can be found in appendix.} 
\begin{tabular}{@{}l|lll|lll|lll@{}}
\toprule
         & \multicolumn{3}{c|}{ResNet-34 (74.4)} & \multicolumn{3}{c|}{VGG-13 (66.1)} & \multicolumn{3}{c}{MobileNet-V1 (67.4)} \\ \midrule
\%pruned      & 25\%     & 50\%     & 75\%     & 25\%    & 50\%    & 75\%    & 25\%      & 50\%      & 75\%     \\ \midrule
BN       & 74.0     & 73.4     & 71.5     & 66.7   & 66.0   & 65.7   & 67.5     & \textbf{68.2}      & 65.6     \\ 
Fisher   & 74.1     & 73.6     & 72.2     & 66.7   & 66.2    & 65.9    & 67.8     & 68.0     & \textbf{66.0}     \\
OrthoReg & \textbf{77.0}     & \textbf{76.7}     & \textbf{74.4}     & \textbf{71.2}   & \textbf{71.3}    & \textbf{67.4}   & \textbf{68.0}     & 67.7     & 65.9     \\ \bottomrule
\end{tabular}
\end{table}

\subsection{Early-Bird Tickets: Improved Trainability and Robustness}
\label{sec:ebt}
You et al.~\cite{ebt} recently showed that after a minimal amount of training (e.g., 10\% of total epochs), a network becomes amenable to pruning and the pruned networks, dubbed Early-Bird Tickets, achieve performance similar to or better than the original networks after complete training. This idea was inspired from recent empirical work by Achille et al.~\cite{critical-learning}, who show that until the aforementioned minimal training period, the network's Fisher information with the output continuously increases, beginning to decrease after that. You et al.\ thus argue that the connectivity of a network is established early on, and therefore less important filters can be removed after minimal training. As the process of extracting Early-Bird Tickets is single-shot, reliable estimation of the importance of a group of filters is paramount. You et al.\ use BatchNorm scale parameters as the importance of a filter~\cite{netslim}. However, \autoref{eqn:ind_imp} is actually a weighted estimate of the Fisher information of a filter~\cite{fisher}. Therefore, using it as an importance metric, instead of heuristic scale parameters, should better fit with the analysis by Achille et al., thus making OrthoReg well suited for extracting Early-Bird Tickets. 

A more important issue in extracting Early-Bird Tickets is that for extreme pruning regimes (e.g., 75\% pruning), the network inevitably loses important connections as well, thus impacting trainability~\cite{signalprop}. This implies ensuring high quality training dynamics in resulting pruned networks is crucial for enabling extreme pruning regimes. Since OrthoReg improves trainability in pruned networks by enforcing orthonormal filters, it is capable of producing better performing Early-Bird Tickets under extreme pruning as well. To evaluate the same, following the exact training setup used by You et al. (see appendix for details), we train ResNet-34, VGG-13, and MobileNet-V1 models on CIFAR-100 for 9\%, 15\%, and 15\% of the total training time (160 epochs), respectively. We prune these minimally trained models with the BatchNorm scale-based method used by You et al.\ (BN), the Fisher information based method (Fisher), and OrthoReg (\autoref{training-algo} with $N = 1$). After pruning, the networks are trained to completion. As shown in \autoref{tab:EBT}, Early-Bird Tickets extracted using OrthoReg consistently achieve much higher test accuracy than other methods. This shows that by preserving orthonormality, OrthoReg's improved trainability of pruned models enables training of high-accuracy models in extreme pruning regimes as well. For example, even when 75\% of filters are pruned in a single round, OrthoReg-pruned models achieve much higher accuracy than both the original networks and the networks pruned using prior methods. 

\subsection{Iterative Pruning}
\begin{table}[]
\caption{\label{tab:cifar}%
Accuracy (Acc.) and efficiency (Eff.) of pruned networks on CIFAR-100. OrthoReg achieves better accuracy than prior methods, while achieving better efficiency for one-third of the models.}
\begin{tabular}{@{}lllllllllllll@{}}
\toprule
\multicolumn{1}{l|}{} & \multicolumn{2}{c|}{OrthoReg} & \multicolumn{2}{c|}{Fisher} & \multicolumn{2}{c|}{TFO} & \multicolumn{2}{c|}{RDT} & \multicolumn{2}{c|}{L1} & \multicolumn{2}{c}{SFP} \\ \midrule
\multicolumn{1}{l|}{\%p} & Acc. & \multicolumn{1}{l|}{Eff.} & Acc. & \multicolumn{1}{l|}{Eff.} & Acc. & \multicolumn{1}{l|}{Eff.} & Acc. & \multicolumn{1}{l|}{Eff.} & Acc. & \multicolumn{1}{l|}{Eff.} & Acc. & Eff. \\ \midrule

\multicolumn{3}{l}{VGG-13 (66.1)} & & & & & & & & & & \\ \cmidrule(r){1-3}
\multicolumn{1}{l|}{65} & \textbf{67.2} & \multicolumn{1}{l|}{5.7} & 67.0 & \multicolumn{1}{l|}{5.6} & 66.7 & \multicolumn{1}{l|}{4.9} & 65.6 & \multicolumn{1}{l|}{6.0} & 64.9 & \multicolumn{1}{l|}{\textbf{6.8}} & 65.6 & 6.5 \\
\multicolumn{1}{l|}{75} & \textbf{65.4} & \multicolumn{1}{l|}{16.5} & 63.9 & \multicolumn{1}{l|}{17.6} & 63.8 & \multicolumn{1}{l|}{17.4} & 63.3 & \multicolumn{1}{l|}{17.5} & 62.2 & \multicolumn{1}{l|}{\textbf{23.3}} & 62.6 & 17.9 \\ \midrule

\multicolumn{4}{l}{MobileNet V1 (67.6)} & & & & & & & & & \\ \cmidrule(r){1-3}
\multicolumn{1}{l|}{65} & \textbf{67.6} & \multicolumn{1}{l|}{6.0} & 66.7 & \multicolumn{1}{l|}{5.6} & 66.7 & \multicolumn{1}{l|}{5.6} & 64.9 & \multicolumn{1}{l|}{\textbf{6.8}} & 66.8 & \multicolumn{1}{l|}{5.8} & 66.1 & 5.8 \\
\multicolumn{1}{l|}{75} & \textbf{66.2} & \multicolumn{1}{l|}{\textbf{12.1}} & 65.0 & \multicolumn{1}{l|}{11.1} & 65.4 & \multicolumn{1}{l|}{11.0} & 62.3 & \multicolumn{1}{l|}{11.5} & 64.5 & \multicolumn{1}{l|}{11.6} & 65.4 & 11.7 \\ \midrule

\multicolumn{3}{l}{ResNet-34 (72.6)} & & & & & & & & & \\ \cmidrule(r){1-3}
\multicolumn{1}{l|}{65} & \textbf{74.1} & \multicolumn{1}{l|}{\textbf{15.9}} & 73.2 & \multicolumn{1}{l|}{15.1} & 72.8 & \multicolumn{1}{l|}{15.3} & 72.8 & \multicolumn{1}{l|}{11.5} & 72.8 & \multicolumn{1}{l|}{13.6} & 72.6 & 15.6 \\
\multicolumn{1}{l|}{85} & \textbf{73.2} & \multicolumn{1}{l|}{37.6} & 72.1 & \multicolumn{1}{l|}{38.4} & 72.3 & \multicolumn{1}{l|}{40.2} & 72 & \multicolumn{1}{l|}{41.1} & 47.2 & \multicolumn{1}{l|}{\textbf{203}} & 50.0 & 149 \\ \bottomrule
\end{tabular}
\end{table}

We evaluate OrthoReg for iterative pruning on an all-convolutional variant of VGG-13, MobileNet-V1, and ResNet-34. These models represent vanilla convolutional networks (VGG-13); networks specifically designed for low redundancies (MobileNet-V1); and networks with special structural constraints (e.g., skip connections) that make them harder to prune (ResNet-34). The models are trained and evaluated on CIFAR-100 and Tiny-ImageNet. We compare our work with five prior methods--L1~\cite{l1}, SFP~\cite{soft}, RDT~\cite{rdt}, TFO~\cite{tfo}, and Fisher~\cite{fisher, nvidia}. L1 and SFP are magnitude-based pruning strategies. RDT provably achieves optimal compression for deep linear and 1-layer nonlinear networks. Fisher and TFO are, to the best of our knowledge, the SOTA methods. For a fair evaluation, all baselines were re-implemented. At each round of pruning, we remove a fixed percentage of filters that are globally least important. The networks are retrained with a learning rate of $10^{-3}$ for 100 epochs with a drop of $10$ at the $40^{\text{th}}$ and $80^{\text{th}}$ epochs. All networks are trained using Adam with weight decay ($\lambda=0.0005$). We do not use weight decay with OrthoReg because the regularizer enforces unit norm on filters, while weight decay tries to decrease their norm. For OrthoReg, 2 rounds of pruning are performed, while 5 rounds are used for all other methods. Note that 1 round is generally sufficient for preserving performance, if using OrthoReg. However, since all methods improve with more rounds, for a fair comparison, we used at least 2 rounds for OrthoReg as well. We use $\text{CR}$ to denote the compression ratio achieved by pruning the original model, $\text{\#FLOPs}_{prun.}$ to denote the number of FLOPs in the pruned model, and $\text{\#FLOPs}_{orig.}$ to denote the number of FLOPs in the original model. Due to space constraints, we report a single computational efficiency metric, defined as $\text{Eff.} = \text{CR} \times \left(1 - \text{\#FLOPs}_{prun.} / \text{\#FLOPs}_{orig.}\right)$. Increase in compression ratio and reduction in FLOPs result in a higher $\text{Eff.}$ value. Further details on the pruning/training setup, CR, and FLOP reductions can be found in appendix.

Results for CIFAR-100 and Tiny-ImageNet are reported in \autoref{tab:cifar} and \autoref{tab:imagenet}, respectively. The original models' accuracies are appended to their names, in parentheses. We report test accuracy (Acc.) and computational efficiency (Eff.) of pruned models for varying levels of pruning (\%p). At any given amount of pruning, OrthoReg consistently achieves higher accuracy than all other methods and, unlike prior work, is able to prune large portions of a network per round without loss in accuracy. 

\begin{table}[]
\caption{\label{tab:imagenet}%
Accuracy (Acc.) and efficiency (Eff.) of pruned networks on Tiny-ImageNet. OrthoReg achieves better accuracy than prior methods, while achieving better efficiency for half the models.}
\begin{tabular}{@{}lllllllllllll@{}}
\toprule
\multicolumn{1}{l|}{} & \multicolumn{2}{c|}{OrthoReg} & \multicolumn{2}{c|}{Fisher} & \multicolumn{2}{c|}{TFO} & \multicolumn{2}{c|}{RDT} & \multicolumn{2}{c|}{L1} & \multicolumn{2}{c}{SFP} \\ \midrule
\multicolumn{1}{l|}{\%p} & Acc. & \multicolumn{1}{l|}{Eff.} & Acc. & \multicolumn{1}{l|}{Eff.} & Acc. & \multicolumn{1}{l|}{Eff.} & Acc. & \multicolumn{1}{l|}{Eff.} & Acc. & \multicolumn{1}{l|}{Eff.} & Acc. & Eff. \\ \midrule
\multicolumn{3}{l}{VGG-13 (49.5)} & & & & & & & & & & \\ \cmidrule(r){1-3}
\multicolumn{1}{l|}{70} & \textbf{49.4} & \multicolumn{1}{l|}{10.6} & 47.8 & \multicolumn{1}{l|}{10.5} & 48.2 & \multicolumn{1}{l|}{10.4} & 48.9 & \multicolumn{1}{l|}{9.8} & 47.9 & \multicolumn{1}{l|}{11.3} & 47.5 & \textbf{11.4} \\
\multicolumn{1}{l|}{80} & \textbf{47.2} & \multicolumn{1}{l|}{22} & 46.4 & \multicolumn{1}{l|}{22.4} & 45.9 & \multicolumn{1}{l|}{21.9} & 42.4 & \multicolumn{1}{l|}{20.7} & 46.2 & \multicolumn{1}{l|}{23.2} & 45.7 & \textbf{23.8} \\ \midrule
\multicolumn{4}{l}{MobileNet-V1 (46.1)} & & & & & & & & & \\ \cmidrule(r){1-3}
\multicolumn{1}{l|}{65} & \textbf{49.5} & \multicolumn{1}{l|}{\textbf{6.5}} & 48.2 & \multicolumn{1}{l|}{5.9} & 47.5 & \multicolumn{1}{l|}{5.9} & 46.6 & \multicolumn{1}{l|}{6.3} & 48.1 & \multicolumn{1}{l|}{5.9} & 48.0 & 5.9 \\
\multicolumn{1}{l|}{82} & 45.8 & \multicolumn{1}{l|}{\textbf{23.2}} & 45.3 & \multicolumn{1}{l|}{23} & 45.4 & \multicolumn{1}{l|}{21.8} & 44 & \multicolumn{1}{l|}{21.8} & 44.6 & \multicolumn{1}{l|}{22.4} & \textbf{46.0} & 22.7 \\ \midrule
\multicolumn{4}{l}{ResNet-34 (52.4)} & & & & & & & & & \\ \cmidrule(r){1-3}
\multicolumn{1}{l|}{60} & \textbf{54.7} & \multicolumn{1}{l|}{\textbf{13.1}} & 52.7 & \multicolumn{1}{l|}{9.9} & 53.4 & \multicolumn{1}{l|}{9.8} & 52.3 & \multicolumn{1}{l|}{5.6} & 52.2 & \multicolumn{1}{l|}{5.3} & 52.5 & 6.2 \\
\multicolumn{1}{l|}{80} & \textbf{54.1} & \multicolumn{1}{l|}{45.6} & 52.7 & \multicolumn{1}{l|}{44.6} & 52.5 & \multicolumn{1}{l|}{46.3} & 52 & \multicolumn{1}{l|}{40.9} & 47.1 & \multicolumn{1}{l|}{\textbf{90.7}} & 48.6 & 86.4 \\ \bottomrule
\end{tabular}
\end{table}

In general, we make the following observations:
\begin{enumerate}[leftmargin=*]
\item At the same amount of pruning, for both iterative pruning and Early-Bird Tickets, OrthoReg produces significant improvements for models with high redundancies, i.e., VGG-13 and ResNet-34. For models with low-redundancies, i.e,  MobileNet-V1, OrthoReg outperforms other methods until 60--70\% pruning, but at extreme levels of pruning (i.e., > 75\%), OrthoReg's improvements are minimal. This shows OrthoReg and prior work are equally competitive on low-redundancy models, further supporting our claim that correlated filters in overparameterized models influence importance estimates in prior work.

\item For OrthoReg-pruned ResNet-34 models, we found several residual blocks were completely removed without loss in accuracy--e.g., the original architecture, which has $(3, 4, 6, 4)$ residual blocks in 4 layers, was reduced\footnote{Reduce here means only 1 filter remains in the residual block. This constraint ensures entire layers are not removed. For further details, see appendix.} to $(3, 3, 1, 1)$ blocks. These results support the hypothesis of Greff et al., who claim that ResNets are unrolled iterative estimators that refine features using several blocks in a layer~\cite{iter_resnet}. In particular, if features with necessary information have been extracted already, remaining feature refinement blocks are unneeded. As the signal can flow uninterruptedly using skip connections, OrthoReg chooses to remove any unneeded blocks.

\item The computational efficiency of OrthoReg is better than other methods for half the experiments. At large amounts of pruning, magnitude-based methods witness a significant loss in accuracy, while achieving better computational efficiency. This shows magnitude-based strategies do not estimate importance reliably. Moreover, when similar accuracy models are compared, OrthoReg achieves higher efficiency than all other methods. Thus, OrthoReg's reliable importance estimates allow a user to properly trade-off less important filters for improved computational efficiency. 
\end{enumerate}

\section{Conclusion}
In this work, we show that enforcing orthonormality on a network's filters significantly improves its pruned networks' performance over SOTA methods. For this purpose, we design OrthoReg, a regularization strategy that produces better performing pruned models than the best existing approaches and allows pruning of several filters per round. These improvements are achieved through more reliable importance estimates, via orthogonality, and better trainability of pruned models, via orthonormality. Overall, these benefits make OrthoReg robust to large amounts of pruning, resulting in better-performing Early-Bird Tickets and iteratively pruned models than prior work. 

{\small
\bibliographystyle{ieee_fullname}
\bibliography{ms}

\begin{thebibliography}{10}\itemsep=-1pt

\bibitem{critical-learning}
A. {Achille}, M. {Rovere}, and S. {Soatto}.
\newblock {Critical Learning Periods in Deep Networks}.
\newblock In {\em Proc.\ Int.\ Conf.\ on Learning Representations}, 2019.

\bibitem{shrinkbench}
D. {Blalock}, J. {Ortiz}, J. {Frankle}, and J. {Guttag}.
\newblock {What is the State of Neural Network Pruning?}
\newblock In {\em Proc.\ Conf.\ on Machine Learning and Systems}, 2020.

\bibitem{rdt}
W. {Gao}, Y. {Liu}, C. {Wang}, and S. {Oh}.
\newblock {Rate Distortion For Model Compression: From Theory To Practice}.
\newblock In {\em Proc.\ Int.\ Conf.\ on Machine Learning}, 2019.

\bibitem{iter_resnet}
K. {Greff}, R. {Srivastava}, and J. {Schmidhuber}.
\newblock {Highway and Residual Networks Learn Unrolled Iterative Estimation}.
\newblock In {\em Proc.\ Int.\ Conf.\ on Learning Representations}, 2017.

\bibitem{eie}
S. {Han}, X. {Liu}, H. {Mao}, J. {Pu}, A. {Pedram}, M. {Horowitz}, and W.~J.
  {Dally}.
\newblock {EIE: Efficient Inference Engine on Compressed Deep Neural Network}.
\newblock In {\em Proc.\ Int.\ Symp.\ on Computer Architecture}, 2016.

\bibitem{han}
S. {Han}, J. {Pool}, J. {Tran}, and W.~J. {Dally}.
\newblock {Learning Both Weights and Connections For Efficient Neural
  Networks}.
\newblock In {\em Advances in Neural Information Processing Systems}, 2016.

\bibitem{obs}
B. {Hassibi} and D.~G. {Stork}.
\newblock {Second Order Derivatives For Network Pruning: Optimal Brain
  Surgeon}.
\newblock In {\em Advances in Neural Information Processing Systems}, 1993.

\bibitem{soft}
Y. {He}, G. {Kang}, X. {Dong}, Y. {Fu}, and Y. {Yang}.
\newblock {Soft Filter Pruning For Accelerating Deep Convolutional Neural
  Networks}.
\newblock In {\em Proc.\ Int.\ Joint Conf.\ on Artificial Intelligence}, 2018.

\bibitem{parcorr}
{J. C.} Helton, {J. D.} Johnson, {C. J.} Sallaberry, and {C. B.} Storlie.
\newblock {Survey of Sampling-Based Methods For Uncertainty and Sensitivity
  Analysis}.
\newblock {\em Reliability Engineering and System Safety}, 2006.

\bibitem{cka}
S. {Kornblith}, M. {Norouzi}, H. {Lee}, and G. {Hinton}.
\newblock {Similarity of Neural Network Representations Revisited}.
\newblock In {\em Proc.\ Int.\ Conf.\ on Machine Learning}, 2019.

\bibitem{obd}
Y. {LeCun}, J.~S. {Denker}, and S.~A. {Solla}.
\newblock {Optimal Brain Damage}.
\newblock In {\em Advances in Neural Information Processing Systems}, 1990.

\bibitem{signalprop}
N. {Lee}, T. {Ajanthan}, S. {Gould}, and P. {Torr}.
\newblock {A Signal Propagation Perspective For Pruning Neural Networks at
  Initialization}.
\newblock In {\em Proc.\ Int.\ Conf.\ on Learning Representations}, 2020.

\bibitem{snip}
N. {Lee}, T. {Ajanthan}, and P. {Torr}.
\newblock {SNIP}: {Single}-{Shot} {Network} {Pruning} {Based} on {Connection}
  {Sensitivity}.
\newblock In {\em Proc.\ Int.\ Conf.\ on Learning Representations}, 2019.

\bibitem{l1}
H. {Li}, A. {Kadav}, I. {Durdanovic}, H. {Samet}, and H.~P. {Graf}.
\newblock {Pruning Filters For Efficient ConvNets}.
\newblock In {\em Proc.\ Int.\ Conf.\ on Learning Representations}, 2017.

\bibitem{netslim}
Z. {Liu}, J. {Li}, Z. {Shen}, G. {Huang}, S. {Yan}, and C. {Zhang}.
\newblock {Learning Efficient Convolutional Networks Through Network Slimming}.
\newblock In {\em Proc.\ Int.\ Conf.\ on Computer Vision}, 2017.

\bibitem{nvidia}
P. {Molchanov}, A. {Mallya}, S. {Tyree}, I. {Frosio}, and J. {Kautz}.
\newblock {Importance Estimation For Neural Network Pruning}.
\newblock In {\em Proc.\ Int.\ Conf.\ on Computer Vision and Pattern
  Recognition}, 2019.

\bibitem{tfo}
P. {Molchanov}, S. {Tyree}, T. {Karras}, T. {Aila}, and J. {Kautz}.
\newblock {Pruning Convolutional Neural Networks For Resource Efficient
  Inference}.
\newblock In {\em Proc.\ Int.\ Conf.\ on Learning Representations}, 2017.

\bibitem{pwcca}
A.~S. {Morcos}, M. {Raghu}, and S. {Bengio}.
\newblock {Insights on Representational Similarity in Neural Networks with
  Canonical Correlation}.
\newblock In {\em Advances in Neural Information Processing Systems}, 2018.

\bibitem{svcca}
M. {Raghu}, J. {Gilmer}, J. {Yosinski}, and J. {Sohl-Dickstein}.
\newblock {SVCCA: Singular Vector Canonical Correlation Analysis For Deep
  Learning Dynamics and Interpretability}.
\newblock In {\em Advances in Neural Information Processing Systems}, 2017.

\bibitem{saxe}
A.~M. {Saxe}, J.~L. {McClelland}, and S. {Ganguli}.
\newblock {Exact Solutions to the Nonlinear Dynamics of Learning in Deep Linear
  Neural Networks}.
\newblock In {\em Advances in Neural Information Processing Systems}, 2014.

\bibitem{fisher}
L. {Theis}, I. {Korshunova}, A. {Tejani}, and F. {H}usz{\'{a}}r.
\newblock {Faster Gaze Prediction with Dense Networks and Fisher Pruning}.
\newblock In {\em Proc.\ Euro. Conf.\ on Computer Vision}, 2018.

\bibitem{ebt}
H. {You}, C. {Li}, P. {Xu}, Y. {Fu}, Y. {Wang}, X. {Chen}, R.~G. {Baraniuk}, Z.
  {Wang}, and Y. {Lin}.
\newblock {Drawing Early-Bird Tickets: Towards More Efficient Training of Deep
  Networks}.
\newblock In {\em Proc.\ Int.\ Conf.\ on Learning Representations}, 2020.

\end{thebibliography}
}

\newpage
\appendix
{\bf \Large Appendix}
\section{Organization}
The appendix is organized as follows:
\begin{itemize}
  \item \autoref{sec:setup}: Experimental setup details
  \item \autoref{sec:train}: Training details
  \item \autoref{sec:pruning}: Design decisions in network pruning  
  \item \autoref{sec:iterative}: Detailed results for Iterative Pruning
  \item \autoref{sec:ebt_appendix}: Detailed results for Early-Bird Tickets
\end{itemize}

\section{Setup Details}
\label{sec:setup}

\subsection{Models}
\begin{enumerate}

\item VGG-13: The model was converted into a fully convolutional version by replacing its original fully connected layers with convolutional layers.

\item MobileNet-V1: MobileNets are low-redundancy models designed for improving accessibility of deep learning techniques on edge devices. 

\item ResNet-34: ResNet-like architectures have proven useful in most modern applications. However, ResNets are difficult to prune, as the output and the input of a residual block must have the same number of channels. This constraint is enforced by the use of skip connections, which add the input of a residual block to its output.

\end{enumerate}

\subsection{Datasets}
\begin{enumerate}

\item CIFAR-100: available at \url{https://www.cs.toronto.edu/~kriz/cifar.html}.

\item Tiny-ImageNet: available at \url{https://tiny-imagenet.herokuapp.com/}.

\end{enumerate}

\subsection{Preprocessing}
The datasets were normalized using following channel-wise mean and standard deviation:
\begin{enumerate}

\item Tiny-ImageNet: mean = [0.425, 0.394, 0.349], standard deviation = [0.298, 0.287, 0.286]

\item CIFAR-100\footnote{The use of synthetic mean and standard deviation for CIFAR-100 was based on the official release of PyTorch tutorial code for training CIFAR classifiers (see \url{https://pytorch.org/tutorials/beginner/blitz/cifar10_tutorial.html}).}: mean = [0.5, 0.5, 0.5], standard deviation = [0.5, 0.5, 0.5]

\end{enumerate}

\subsection{Data Augmentation}
The training dataset was augmented as follows:
\begin{enumerate}

\item Random crop of square size 32 with padding of 4 pixels

\item Random horizontal flips

\end{enumerate}

\subsection{Equipment}
\begin{enumerate}

\item All experiments were performed using a single NVIDIA GeForce RTX 2070. 

\end{enumerate}

\section{Training details}
\label{sec:train}

\subsection{Iterative Pruning}
\subsubsection{Training Base Models}
\begin{enumerate}

\item Optimizer: Adam (betas=$(0.9, 0.999)$)

\item Weight decay: $0.0005$

\subitem For OrthoReg, weight decay is set to 0 

\item Learning rate schedule: $(0.001, 0.0001, 0.00001)$
\subitem For training with Orthonormality Regularization, learning rate schedule is: $(0.001, 0.0001, 0.00001, 0.000001)$

\item Number of epochs for each learning rate: $(80, 40, 40)$
\subitem For training with Orthonormality Regularization, number of epochs are $(10, 10, 10, 5)$
\item Batch Size: $128$

\item $\lambda$ for OrthoReg (See Equation 4 of main paper): 

\subitem VGG-13 and ResNet-34: $0.01$

\subitem  MobileNet-V1: $0.001$

\end{enumerate}
Few epochs at lower learning rate substantially improve orthonormality of models. We thus use 5 epochs at a learning rate of $0.000001$ for OrthoReg models. For deciding $\lambda$, we tried 3 values--$0.1,\,0.01,$ and $0.001$. For ResNet-34 and VGG-13, all values performed well, but $0.01$ achieved orthogonality much faster. For MobileNet-V1, we found that $0.1$ resulted in training instability. While both $0.01$ and $0.001$ worked well, we preferred $0.001$ because it resulted in better validation accuracy.

\subsubsection{Training Pruned Models}
The same settings as for a base model are used, except for the following:
\begin{enumerate}
  \item Number of epochs for each learning rate: $(40, 40, 20)$
\end{enumerate}
Note that for training a pruned model with Orthonormality Regularization, weight decay is set to 0.

\subsection{Early-Bird Tickets}
For a fair evaluation, the same training settings were used for Early-Bird Ticket experiments as proposed by the authors of the original work~\cite{ebt} proposed. In particular:
\subsubsection{Training Base Models}
\begin{enumerate}
  \item Optimizer: SGD (momentum=$0.9$)
  \item weight decay: $0.0001$
    \subitem For training with Orthonormality Regularization, weight decay is set to 0.
  \item Learning rate schedule: $(0.1)$
    \subitem For training with Orthonormality Regularization, the schedule used is: $(0.1, 0.01)$
  \item Number of epochs for each learning rate: 
    \subitem ResNet-34: $(15)$; with Orthonormality Regularization: $(14, 1)$
    \subitem VGG-13: $(25)$; with Orthonormality Regularization: $(24, 1)$
    \subitem MobileNet-V1: $(25)$; with Orthonormality Regularization: $(24, 1)$
  \item Batch Size: $128$
  \item $\lambda$ for OrthoReg (See Equation 6 of main paper): 
    \subitem VGG-13 and ResNet-34: $0.01$
    \subitem MobileNet-V1: $0.001$
\end{enumerate}

\subsubsection{Training Early-Bird Tickets (Pruned Models)}
The same settings as for a base model are used, except for the following:
\begin{enumerate}
  \item Learning rate schedule: $(0.1, 0.01, 0.001)$
  \item Number of epochs for each learning rate: $(60, 40, 40)$
\end{enumerate}

\subsection{Regularizing Filters}
As detailed in the main paper, training with Orthonormality Regularization helps improve training dynamics of pruned models by virtue of Layerwise Dynamical Isometry. In particular, Layerwise Dynamical Isometry requires singular values of network's weight matrices to be concentrated around 1. For a layer with weight matrix $\mathbf{W} \in R^{hwc_{in} \times c_{out}}$, i.e., a layer with $c_{out}$ filters with height $h$, width $w$, and $c_{in}$ channels, the matrix $\mathbf{W}^{T}\mathbf{W}$ can be made orthonormal only if $c_{out} > hwc_{in}$, i.e., if the number of filters in a layer are greater than the number of parameters in a single filter in that layer. Note that making $\mathbf{W}^{T}\mathbf{W}$ orthonormal is necessary for retrieving reliable importance estimates. This condition is generally true because number of filters in a layer are usually much less than the number of parameters in a single filter. However, for some special cases, such as the first convolutional layer of a model or early pointwise convolutional layers, number of parameters in a single filter may be less than the number of filters in that layer. In such cases, we enforce orthonormality on $\mathbf{W}\mathbf{W}^{T}$. This ensures that at least singular values of a layer's weight matrix concentrate around 1, thereby improving training dynamics of pruned models.

\section{Design Decisions in Network Pruning}
\label{sec:pruning}

\subsection{Deciding Pruning Ratios}
\label{sec:ratios}
If $p\%$ filters are to be pruned in $n$ rounds, then we set the pruning ratios for each round as follows
\begin{align*}
&p_1 = \frac{\frac{p}{n}}{(1-p) + \frac{p}{n}},\\
&p_2 = \frac{\frac{p}{n}}{(1-p) + \frac{2p}{n}},\\
&\dots\\
&p_n = \frac{\frac{p}{n}}{(1-p) + \frac{np}{n}}.
\end{align*}
When remaining proportions at each round of pruning are multiplied, the net remaining proportion is
\begin{align*}
\text{\% pruned} &= 1 - (p_1 \times p_2 \times \dots \times p_n) \\
&= 1 - \left(\frac{1-p}{(1-p) + \frac{p}{n}} \times \frac{(1-p) + \frac{p}{n}}{(1-p) + \frac{2p}{n}} \times \dots \frac{(1-p) + \frac{(n-1)p}{n}}{(1-p) + \frac{np}{n}}\right)\\
&= 1 - (1-p) = p
\end{align*}
These ratios prune the desired amount of filters, i.e., $p\%$ filters, in $n$ rounds. More importantly, these ratios ensure more pruning happens first, when redundancies are high and network is more amenable to pruning, while pruning less in later rounds.

In our code, we also provide an option to use manual pruning ratios (see --thresholds in README.md).

\subsection{Pruning Methodology for Different Models}
For any convolutional layer, the following pruning method is used: if a filter is deemed unimportant, it is removed. Also, the channels in next layer's filters corresponding to the output of the pruned filter are removed. However, this methodology only works for vanilla convolutional networks (e.g., VGG) that do not have skip connections or depth-separable filters. We prune the models with skip connections (ResNets) and depth-separable filters (MobileNet-V1) as described below.

\subsubsection{Skip Connections: Pruning ResNets}
\begin{figure*}
\centering
\includegraphics[width=\linewidth]{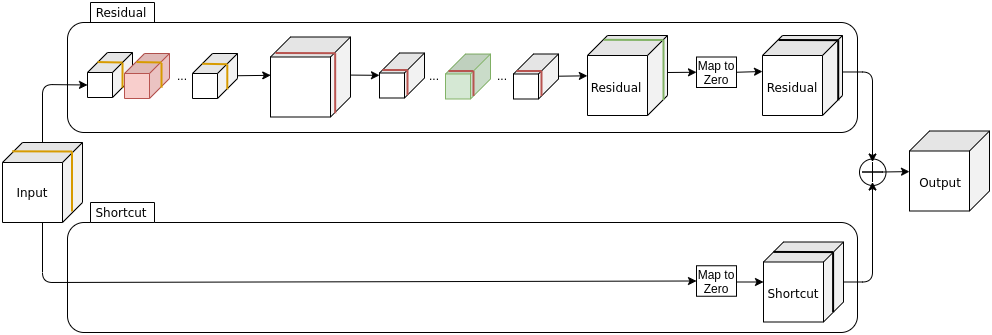}
\caption{ResNet pruning. The colored filters (marked red and green) are pruned and their corresponding channels are thus removed as well. Skip connections require the output of a Residual block to have the same number of channels as the input to the block because of the addition operation at the end. To deal with this, we map the pruned channel (marked green) in a residual block's output to an all zero channel (marked black) before the addition operation. The same is done for the channels pruned at the input (changes from yellow to black), corresponding to which channels (marked yellow) in the filters of first layer of Residual block are removed and the channel (marked black) in the shortcut has an all zero channel mapped at its location.}
\label{fig:res_pruning}
\end{figure*}

Skip connections require the output of a Residual block to have the same number of channels as the input to the block because of the addition operation at the end. Prior works generally find the importance of channels in the input to a Residual block and remove unimportant channels from the input. Correspondingly channels in the Residual output, and consequently filters that generate those channels, are pruned. The reasoning for using this strategy is that channels in the Residual block output are not as important as the channels in the shortcut path, for they are the essential reason for which ResNets were envisaged. 

In contrast to this, we prune unimportant channels in the input and output of the Residual block independently, without pruning their corresponding channels in the output and input, respectively (see \autoref{fig:res_pruning}). To deal with different dimensionality generated as a consequence of this, we use \emph{map to zero} blocks that remember which indices in the channel dimension have been pruned. During a forward pass, the corresponding locations are filled with all-zero channels. This results in exactly the same dimensions as the original Residual output block would have had, however since the corresponding filters to these all-zero channels are removed, significant improvements are observed in compression ratio and reduction of FLOPs. Recall that a convolution is much more expensive than mere addition of 3-dimensional tensors, which is why this strategy results in much better efficiency than prior methods. 

\subsubsection{Depth-separable filters: Pruning MobileNet-V1}
\begin{figure}
\centering
\includegraphics[width=\linewidth]{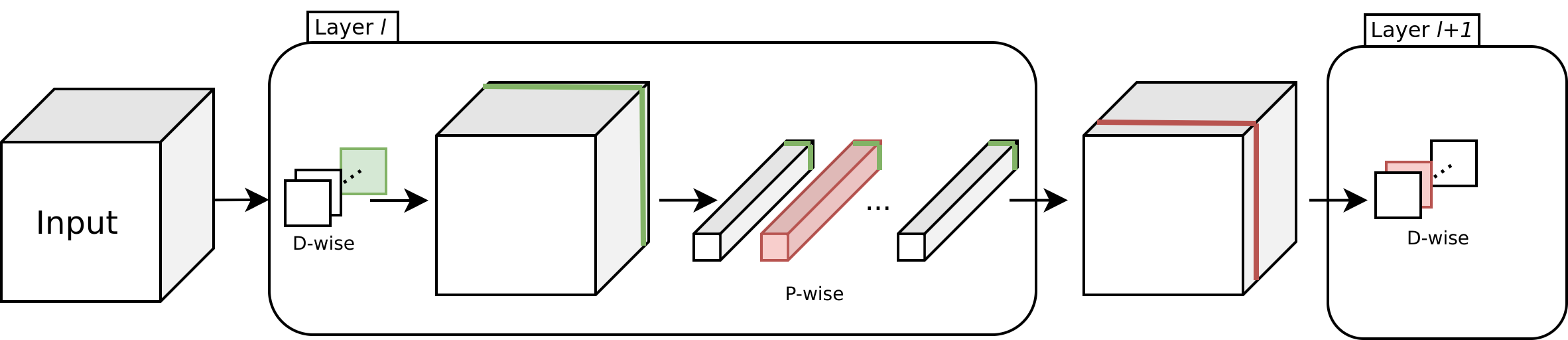}
\caption{MobileNet-V1 pruning. If the red pointwise (P-wise) filter is unimportant and can be pruned, as a consequence of which the corresponding red depthwise (D-wise) filter also has to be pruned. Similarly, if the green depthwise filter is unimportant, it can be pruned and consequently the channels from the pointwise filters have to be removed.}
\label{fig:mobile_pruning}
\end{figure}

MobileNet-V1 has depth-separable filters that use $M$ depthwise filters of dimensions $3\times3\times1$ to process an input with $M$ channels (see \autoref{fig:mobile_pruning}). Each filter processes its corresponding channel, resulting in an output with $M$ channels as well. This output is processed by $N$ pointwise filters of dimensions $1\times1\times N$ filters, mixing information from different channels and producing an output with $N$ channels. If a pointwise filter in layer $l$ is pruned first, its output loses a channel, and thus the corresponding depthwise filter in layer $l+1$ has to be removed. Similarly, if a depthwise filter in layer $l+1$ is removed, its corresponding pointwise filter in layer $l$ has to be removed because there is no depthwise filter to process its output. This results in a constrained pruning structure.  Since pointwise filters aggregate information and result in the vital outputs of a layer, we give higher priority to them and use them to guide the pruning process for depthwise filters--i.e., if a pointwise filter is pruned, its corresponding depthwise filter in the next layer is pruned as well.

\subsection{Constraining Number of Pruned Filters}
Pruning globally can result in removal of an entire layer if all its filters are deemed to be least important in the network. This generally happens with pruning large numbers of neurons in a single round. To deal with this, similar to prior works, we put a constraint that no more than 95\% filters in a layer can be removed. For ResNets, the constraint is relaxed to pruning until only 1 filter remains. The 1 filter constraint is necessary to maintain the architecture.

\section{Detailed Results: Iterative Pruning}
\label{sec:iterative}
Detailed results on iterative pruning are provided in the follow tables. The new information is the exact compression ratio and percentage reduction in FLOPs. We first report the test accuracy and GFLOPs in the base models that were pruned using different strategies. The accuracies are provided for both the original model and their regularized counterparts. We note here that the regularized counterparts do have slightly ($<$1\%) better accuracy in several cases, but after just minimal pruning, the regularized counterparts rarely ever achieved better accuracy than the unregularized models. Usually, the regularized model had slightly worse performance. We report the accuracy of pruned models trained with and without the Orthonormality Regularization in \autoref{tab:orthoreg}.

The results for different pruning techniques are provided in \autoref{tab:orthoreg} (OrthoReg), \autoref{tab:fisher} (Fisher pruning~\cite{fisher, nvidia}), \autoref{tab:TFO} (TFO pruning~\cite{tfo}), \autoref{tab:RDT} (RDT pruning~\cite{rdt}), \autoref{tab:L1} (L1 norm based pruning~\cite{l1}), and \autoref{tab:SFP} (SFP pruning~\cite{soft}). The aforementioned technique in \autoref{sec:ratios} was used for deciding the pruning ratios per round. We used 2 rounds for OrthoReg and 5 rounds for all other methods. We note here that increasing the number of rounds reduced the gap between iterative pruning results for OrthoReg and Fisher pruning, further supporting the conclusions in our main paper that large pruning ratios are not suitable for Fisher pruning.

\begin{table}[H]
\caption{\label{tab:base}%
\textbf{Base models:} Test accuracy and GFLOPs of base models for both the Orthonormality Regularized (Reg.) and Unregularized (Unreg.) versions.}
\centering
\begin{tabular}{@{}c|c|c|c@{}}
\toprule
\textbf{Tiny-ImageNet}     & \textbf{Accuracy (Unreg.)} & \textbf{Accuracy (Reg.)} & \textbf{GFLOPs} \\ \midrule
VGG-13        & 49.5              & 49.9            & 0.229  \\
MobileNet-V1  & 46.1              & 46.7            & 0.047  \\
ResNet-34     & 52.4              & 53.2            & 1.161  \\ \midrule
\textbf{CIFAR-100}     & \textbf{Accuracy (Unreg.)} & \textbf{Accuracy (Reg.)} & \textbf{GFLOPs} \\ \midrule
VGG-13        & 66.1              & 66.6            & 0.229  \\
MobileNet-V1  & 67.6              & 67.5            & 0.047  \\
ResNet-34     & 72.6              & 73.4            & 1.161  \\ \bottomrule
\end{tabular}
\end{table}

\begin{table}[H]
\caption{\label{tab:orthoreg}%
\textbf{OrthoReg pruning:} iterative pruning results. Reported are \% pruning, accuracy of the unregularized model (Unreg.), the regularized model (Reg.), Compression Ratio, and \% FLOPs reduced.}
\centering
\begin{tabular}{@{}c|c|c|c|c|c@{}}
\toprule
\textbf{Tiny-ImageNet} & \textbf{\% pruning} & \textbf{Accuracy}          & \textbf{Accuracy}        & \textbf{Compression}   &   \textbf{\% FLOPs}         \\ 
              &            & \textbf{(Unreg.)}          & \textbf{(Reg.)}          & \textbf{Ratio}             & \textbf{reduced}           \\ \midrule
VGG-13        & 70         & 49.4              & 48.1            & 14.6              & 72.4              \\
              & 80         & 47.2              & 46.5            & 27.3              & 80.7              \\ \midrule
MobileNet-V1  & 65         & 49.5              & 46.1            & 9.3               & 68.1              \\
              & 82         & 45.8              & 43.6            & 27.6              & 84.2              \\ \midrule
ResNet-34     & 60         & 54.7              & 54.5            & 15.7              & 83.5              \\
              & 80         & 54.1              & 53.9            & 52.7              & 86.7              \\ \midrule
\textbf{CIFAR-100} & \textbf{\% pruning} & \textbf{Accuracy}          & \textbf{Accuracy}        & \textbf{Compression}   &   \textbf{\% FLOPs}         \\ 
              &            & \textbf{(Unreg.)}          & \textbf{(Reg.)}          & \textbf{Ratio}             & \textbf{reduced}           \\ \midrule
VGG-13        & 65         & 67.2              & 66              & 9.3               & 60.7              \\
              & 75         & 65.4              & 64.6            & 22.2              & 74.0              \\ \midrule
MobileNet-V1  & 65         & 67.6              & 66.3            & 7.6               & 62.0              \\
              & 75         & 66.2              & 64.5            & 15.8              & 76.1              \\ \midrule
ResNet-34     & 65         & 74.1              & 73.55           & 18.4              & 83.5              \\
              & 85         & 73.2              & 72.94           & 42.6              & 88.0              \\ \bottomrule
\end{tabular}
\end{table}

\begin{table}[H]
\caption{\label{tab:fisher}%
\textbf{Fisher pruning:} iterative pruning results. Reported are \% pruning, accuracy of pruned model, accuracy of the regularized model (Accuracy (Reg.)), Compression Ratio, and \% FLOPs reduced.}
\centering
\begin{tabular}{@{}c|c|c|c|c@{}}
\toprule
\textbf{Tiny-ImageNet} & \textbf{\% pruning} & \textbf{Accuracy} & \textbf{Compression Ratio} & \textbf{\% FLOPs reduced} \\ \midrule
VGG-13        & 70         & 47.8     & 14.7              & 71.3              \\
              & 80         & 46.4     & 28                & 80.0              \\ \midrule
MobileNet-V1  & 65         & 48.2     & 9.3               & 64                \\
              & 82         & 45.3     & 28.5              & 80.8              \\ \midrule
ResNet-34     & 60         & 52.7     & 12.1              & 82.1              \\
              & 80         & 52.7     & 51                & 87.5              \\ \midrule
\textbf{CIFAR-100} & \textbf{\% pruning} & \textbf{Accuracy} & \textbf{Compression Ratio} & \textbf{\% FLOPs reduced} \\ \midrule
VGG-13        & 65         & 67       & 8.9               & 62.2              \\
              & 75         & 63.9     & 22                & 80.3              \\ \midrule
MobileNet-V1  & 65         & 66.7     & 8.8               & 62.9              \\
              & 75         & 65       & 15.4              & 72.2              \\ \midrule
ResNet-34     & 65         & 73.2     & 18                & 83.9              \\
              & 85         & 72.1     & 43.5              & 88.4              \\ \bottomrule
\end{tabular}
\end{table}

\begin{table}[H]
\caption{\label{tab:TFO}%
\textbf{TFO pruning:} iterative pruning results. Reported are \% pruning, accuracy of pruned model, accuracy of the regularized model (Accuracy (Reg.)), Compression Ratio, and \% FLOPs reduced.}
\centering
\begin{tabular}{@{}c|c|c|c|c@{}}
\toprule
\textbf{Tiny-ImageNet} & \textbf{\% pruning} & \textbf{Accuracy} & \textbf{Compression Ratio} & \textbf{\% FLOPs reduced} \\ \midrule
VGG-13        & 70         & 48.2     & 14.6              & 71.4              \\
              & 80         & 45.9     & 27.3              & 80.4              \\ \midrule
MobileNet-V1  & 65         & 47.5     & 9.3               & 64.3              \\
              & 82         & 45.4     & 28.5              & 81                \\ \midrule
ResNet-34     & 60         & 53.4     & 12                & 82.1              \\
              & 80         & 52.5     & 52.6              & 88                \\ \midrule
\textbf{CIFAR-100} & \textbf{\% pruning} & \textbf{Accuracy} & \textbf{Compression Ratio} & \textbf{\% FLOPs reduced} \\ \midrule
VGG-13        & 65         & 66.7     & 8.6               & 56.6              \\
              & 75         & 63.8     & 22.2              & 78.4              \\ \midrule
MobileNet-V1  & 65         & 66.7     & 8.8               & 63.2              \\
              & 75         & 65.4     & 15.2              & 73                \\ \midrule
ResNet-34     & 65         & 72.8     & 18.9              & 84.4              \\
              & 85         & 72.3     & 45.2              & 89                \\ \bottomrule
\end{tabular}
\end{table}

\begin{table}[H]
\caption{\label{tab:RDT}%
\textbf{RDT pruning:} iterative pruning results. Reported are \% pruning, accuracy of pruned model, accuracy of the regularized model (Accuracy (Reg.)), Compression Ratio, and \% FLOPs reduced.}
\centering
\begin{tabular}{@{}c|c|c|c|c@{}}
\toprule
\textbf{Tiny-ImageNet} & \textbf{\% pruning} & \textbf{Accuracy} & \textbf{Compression Ratio} & \textbf{\% FLOPs reduced} \\ \midrule
VGG-13        & 70         & 48.9     & 14                & 70.1              \\
              & 80         & 42.4     & 26.7              & 77.7              \\ \midrule
MobileNet-V1  & 65         & 46.6     & 9.4               & 69.1              \\
              & 82         & 44       & 24.7              & 88.5              \\ \midrule
ResNet-34     & 60         & 52.3     & 7.2               & 78.3              \\
              & 80         & 52       & 46.6              & 87.8              \\ \midrule
\textbf{CIFAR-100} & \textbf{\% pruning} & \textbf{Accuracy} & \textbf{Compression Ratio} & \textbf{\% FLOPs reduced} \\ \midrule
VGG-13        & 65         & 65.6     & 9.1               & 66.2              \\
              & 75         & 63.3     & 21.3              & 82.2              \\ \midrule
MobileNet-V1  & 65         & 64.9     & 9.2               & 73.9              \\
              & 75         & 62.3     & 14.1              & 81.2              \\ \midrule
ResNet-34     & 65         & 72.8     & 13.8              & 83.1              \\
              & 85         & 72       & 45.9              & 89.5              \\ \bottomrule
\end{tabular}
\end{table}

\begin{table}[H]
\caption{\label{tab:L1}%
\textbf{L1-norm based pruning:} iterative pruning results. Reported are \% pruning, accuracy of pruned model, accuracy of the regularized model (Accuracy (Reg.)), Compression Ratio, and \% FLOPs reduced.}
\centering
\begin{tabular}{@{}c|c|c|c|c@{}}
\toprule
\textbf{Tiny-ImageNet} & \textbf{\% pruning} & \textbf{Accuracy} & \textbf{Compression Ratio} & \textbf{\% FLOPs reduced} \\ \midrule
VGG-13        & 70         & 47.9     & 15.3              & 74                \\
              & 80         & 46.2     & 27.9              & 83.2              \\ \midrule
MobileNet-V1  & 65         & 48.1     & 9.3               & 64.1              \\
              & 82         & 44.6     & 27.3              & 82.2              \\ \midrule
ResNet-34     & 60         & 52.2     & 7                 & 76.5              \\
              & 80         & 47.1     & 100               & 90.7              \\ \midrule
\textbf{CIFAR-100} & \textbf{\% pruning} & \textbf{Accuracy} & \textbf{Compression Ratio} & \textbf{\% FLOPs reduced} \\ \midrule
VGG-13        & 65         & 64.9     & 11.8              & 57.8              \\
              & 75         & 62.2     & 27.1              & 86.3              \\ \midrule
MobileNet-V1  & 65         & 66.8     & 9.1               & 63.6              \\
              & 75         & 64.5     & 15.7              & 74.2              \\ \midrule
ResNet-34     & 65         & 72.8     & 16.4              & 82.9              \\
              & 85         & 47.2     & 212               & 96.2              \\ \bottomrule
\end{tabular}
\end{table}

\begin{table}[H]
\caption{\label{tab:SFP}%
\textbf{SFP pruning:} iterative pruning results. Reported are \% pruning, accuracy of pruned model, accuracy of the regularized model (Accuracy (Reg.)), Compression Ratio, and \% FLOPs reduced.}
\centering
\begin{tabular}{@{}c|c|c|c|c@{}}
\toprule
\textbf{Tiny-ImageNet} & \textbf{\% pruning} & \textbf{Accuracy} & \textbf{Compression Ratio} & \textbf{\% FLOPs reduced} \\ \midrule
VGG-13        & 70         & 47.5     & 15.4              & 74.1              \\
              & 80         & 45.7     & 29                & 82.3              \\ \midrule
MobileNet-V1  & 65         & 48.02    & 9.3               & 64                \\
              & 82         & 46.08    & 28.6              & 79.6              \\ \midrule
ResNet-34     & 60         & 52.54    & 7.9               & 78.5              \\
              & 80         & 48.6     & 96.1              & 89.9              \\ \midrule
\textbf{CIFAR-100} & \textbf{\% pruning} & \textbf{Accuracy} & \textbf{Compression Ratio} & \textbf{\% FLOPs reduced} \\ \midrule
VGG-13        & 65         & 65.6     & 11                & 59.3              \\
              & 75         & 62.6     & 23.2              & 77.2              \\ \midrule
MobileNet-V1  & 65         & 66.1     & 9.1               & 63.6              \\
              & 75         & 65.4     & 15.9              & 73.6              \\ \midrule
ResNet-34     & 65         & 72.6     & 18.6              & 84                \\
              & 85         & 50       & 158               & 94.7              \\ \bottomrule
\end{tabular}
\end{table}

\section{Detailed Results: Early-Bird Tickets}
\label{sec:ebt_appendix}
Detailed results of the Early-Bird Tickets extracted using different pruning methods are provided below. The new information is Compression Ratio and percentage reduction in FLOPs for the compressed models. 

In \autoref{tab:ebt_base}, we report the accuracy of the base, unpruned models after training to completion. The extraction of Early-Bird Tickets happens after minimal amount of training (9\% for ResNet, 15\% for VGG and MobileNet). The numbers in \autoref{tab:ebt_base} provide a useful baseline for evaluating the efficiency and accuracy of compressed models. The test accuracy, compression ratio, and percentage reduction in FLOPs for Early-Bird Tickets extracted are reported in \autoref{tab:ebt_orthoreg} (OrthoReg), \autoref{tab:ebt_fisher} (Fisher pruning~\cite{fisher, nvidia}), and \autoref{tab:ebt_bn} (BN-scale based pruning~\cite{netslim}).

\begin{table}[H]
\centering
\caption{\label{tab:ebt_base}%
\textbf{Fully trained base models:} Test accuracy and GFLOPs of fully trained base models used for extracting Early-Bird Tickets.}
\begin{tabular}{@{}c|c|c@{}}
\toprule
\textbf{Model}        & \textbf{Accuracy}       & \textbf{GFLOPs}  \\ \midrule
ResNet-34     & 74.4              & 1.161  \\ 
VGG-13        & 66.1              & 0.229  \\
MobileNet-V1  & 67.4              & 0.047  \\ \bottomrule
\end{tabular}
\end{table}

\begin{table}[H]
\centering
\caption{\label{tab:ebt_orthoreg}%
\textbf{OrthoReg pruning based Early-Bird Tickets:} Reported are test Accuracy, Compression Ratio, and \% FLOPs reduced.}
\begin{tabular}{@{}c|c|c|c@{}}
\toprule
\textbf{ResNet-34}    & \textbf{Accuracy} & \textbf{Compression Ratio} & \textbf{\% FLOPs Reduced} \\ \midrule
25\%         & 77.0     & 1.5               & 14.9             \\
50\%         & 76.7     & 2.8               & 38.1             \\
75\%         & 74.4     & 9.0               & 66.5             \\ \midrule
\textbf{VGG-13}    & \textbf{Accuracy} & \textbf{Compression Ratio} & \textbf{\% FLOPs Reduced} \\ \midrule
25\%         & 71.2     & 1.9               & 18.5             \\
50\%         & 71.3       & 5.1                 & 35.0             \\
75\%         & 67.4     & 13.4                & 55.8             \\ \midrule
\textbf{MobileNet-V1}    & \textbf{Accuracy} & \textbf{Compression Ratio} & \textbf{\% FLOPs Reduced} \\ \midrule
25\%         & 68.0     & 1.9               & 32.6             \\
50\%         & 67.7     & 4.5               & 58.9             \\
75\%         & 65.9     & 17.1              & 79.7             \\ \bottomrule
\end{tabular}
\end{table}

\begin{table}[H]
\centering
\caption{\label{tab:ebt_fisher}%
\textbf{Fisher pruning based Early-Bird Tickets:} Reported are test Accuracy, Compression Ratio, and \% FLOPs reduced.}
\begin{tabular}{@{}c|c|c|c@{}}
\toprule
\textbf{ResNet-34}    & \textbf{Accuracy} & \textbf{Compression Ratio} & \textbf{\% FLOPs Reduced} \\ \midrule
25\%                              & 74.1                          & 1.4                                   & 30.0                                 \\
50\%                              & 73.6                          & 2.2                                   & 52.0                                 \\
75\%                              & 72.2                          & 5.5                                   & 75.3                                 \\ \midrule
\textbf{VGG-13}    & \textbf{Accuracy} & \textbf{Compression Ratio} & \textbf{\% FLOPs Reduced} \\ \midrule
25\%                              & 66.7                          & 1.7                                   & 19.4                                 \\
50\%                              & 66.2                          & 4.4                                   & 50.5                                 \\
75\%                              & 65.9                          & 19.2                                   & 73.8                                 \\ \midrule
\textbf{MobileNet-V1}    & \textbf{Accuracy} & \textbf{Compression Ratio} & \textbf{\% FLOPs Reduced} \\ \midrule
25\%                              & 67.8                          & 2.0                                   & 28.7                                 \\
50\%                              & 68.0                          & 4.6                                   & 52.1                                 \\
75\%                              & 66.0                          & 15.3                                   & 74.0                                   \\ \bottomrule
\end{tabular}
\end{table}

\begin{table}[H]
\centering
\caption{\label{tab:ebt_bn}%
\textbf{BN-scale pruning based Early-Bird Tickets:} Reported are test Accuracy, Compression Ratio, and \% FLOPs reduced.}
\begin{tabular}{@{}c|c|c|c@{}}
\toprule
\textbf{ResNet-34}    & \textbf{Accuracy} & \textbf{Compression Ratio} & \textbf{\% FLOPs Reduced} \\ \midrule
25\%                              & 74.0                          & 1.2                                    & 26.3                                 \\
50\%                              & 73.4                          & 2.2                                    & 58.4                                 \\
75\%                              & 71.5                          & 10.3                                    & 86.4                                 \\ \midrule
\textbf{VGG-13}    & \textbf{Accuracy} & \textbf{Compression Ratio} & \textbf{\% FLOPs Reduced} \\ \midrule
25\%                              & 66.7                          & 1.6                                    & 34.6                                 \\
50\%                              & 66.0                          & 3.9                                    & 65.9                                 \\
75\%                              & 65.7                          & 17.3                                   & 84.7                                   \\ \midrule
\textbf{MobileNet-V1}    & \textbf{Accuracy} & \textbf{Compression Ratio} & \textbf{\% FLOPs Reduced} \\ \midrule
25\%                              & 67.5                          & 1.9                                    & 34.0                                 \\
50\%                              & 68.2                          & 4.5                                    & 58.8                                 \\
75\%                              & 65.6                          & 12.4                                   & 78.3                                 \\ \bottomrule
\end{tabular}
\end{table}

\end{document}